\documentclass{article}

\pdfoutput=1

\PassOptionsToPackage{numbers, compress}{natbib}

\usepackage[final]{building_damage_recognition}

\usepackage[utf8]{inputenc} 
\usepackage[T1]{fontenc}    
\usepackage{hyperref}       
\usepackage{url}            
\usepackage{booktabs}       
\usepackage{amsfonts}       
\usepackage{microtype}      
\usepackage{float}
\usepackage{subfig}

\usepackage{graphicx}
\graphicspath{ {./images/} }

\title{Building Damage Detection in Satellite Imagery Using Convolutional Neural Networks}

\author{%
  Joseph Z. Xu \\
  Google AI\\
  \texttt{jzxu@google.com}
  \And
  Wenhan Lu \\
  Google AI\\
  \texttt{wenhan@google.com}
  \And
  Zebo Li \\
  Google AI \\
  \texttt{pkugoodspeed@google.com}
  \And
  Pranav Khaitan \\
  Google AI \\
  \texttt{pranavkhaitan@google.com}
  \And
  Valeriya Zaytseva \\
  UN World Food Programme \\
  \texttt{valeriya.zaytseva@wfp.org}
}

\begin{document}

\maketitle

\begin{abstract}
In all types of disasters, from earthquakes to armed conflicts, aid workers need accurate and timely data such as damage to buildings and population displacement to mount an effective response.
Remote sensing provides this data at an unprecedented scale, but extracting operationalizable information from satellite images is slow and labor-intensive.
In this work, we use machine learning to automate the detection of building damage in satellite imagery.
We compare the performance of four different convolutional neural network models in detecting damaged buildings in the 2010 Haiti earthquake.
We also quantify how well the models will generalize to future disasters by training and testing models on different disaster events.
\end{abstract}

\section{Introduction}

At the start of a humanitarian crisis, it is critical for humanitarian agencies to know the locations of affected populations within the first few hours after a disaster in order to facilitate deployment of response activities.
Damaged buildings are often used as a proxy for affected population localization \cite{dellacqua}.
Remote sensing is a powerful tool for identifying damaged buildings due to its wide coverage area and availability of data.
However, humanitarian actors mostly rely on manual digitization of damaged structures, which remains the most reliable method.
Manual digitization is labor-intensive,  requiring trained image analysts, is unsuitable for large areas, and is prone to inconsistencies related to human errors due to fatigue or quality control.
Automating this process would greatly reduce the time required to produce damage assessment reports.

In the last few years, we have seen rapid advances in the field of machine learning for computer vision, particularly with respect to deep neural networks (DNNs) \cite{Goodfellow:2016:DL:3086952}.
DNNs have achieved human-level performance on a variety of computer vision tasks, including object recognition and image segmentation.
These techniques are therefore suitable for automatically extracting information from satellite images.

Researchers have applied machine learning approaches to building damage detection in satellite imagery.
Cooner et al. \cite{cooner} compared the performance of multiple machine learning methods in building damage detection with both pre-event and post-event satellite imagery of the 2010 Haiti earthquake, and found that a feed-forward neural network achieved the lowest error rate of 40$\%$.
Min Ji et al \cite{min_cnn} developed a convolutional network to identify collapsed buildings from post-event satellite imagery of the Haiti earthquake, and reached an overall accuracy of 78.6$\%$.
Duarte et. al. \cite{duarte} combine drone and satellite images of disasters to improve the accuracy of their convolutional networks, with a best reported accuracy of $94.4\%$.
However, to the best of our knowledge, there has not been any work on cross-region transfer learning, i.e. training models on one region while testing them on another.
Cross-region transfer learning is important because the ultimate test of a model is its accuracy when applied to future disasters, which will likely affect regions that the model has not been trained on.

In this paper, we investigate the generalizability of convolutional neural networks (CNNs) in detecting building damage caused by disasters.
We first compare the performance of different CNN architectures on one dataset, and then compare the performance of the best CNN architecture when trained and validated in different transfer learning contexts.
The paper is structured as follows.
Section \ref{sec:s2} describes the data collection pipeline.
Section \ref{sec:s3} compares the performance of four different CNN architectures using Haiti earthquake dataset.
Section \ref{sec:s4} investigates model generalizability and presents results in cross-region transfer learning experiment.
Section \ref{sec:s5} presents our conclusions and future work.

\section{\label{sec:s2}Data Generation Pipeline}
\label{data_collection}

\begin{figure}[t]
\includegraphics[width=1.0\textwidth]{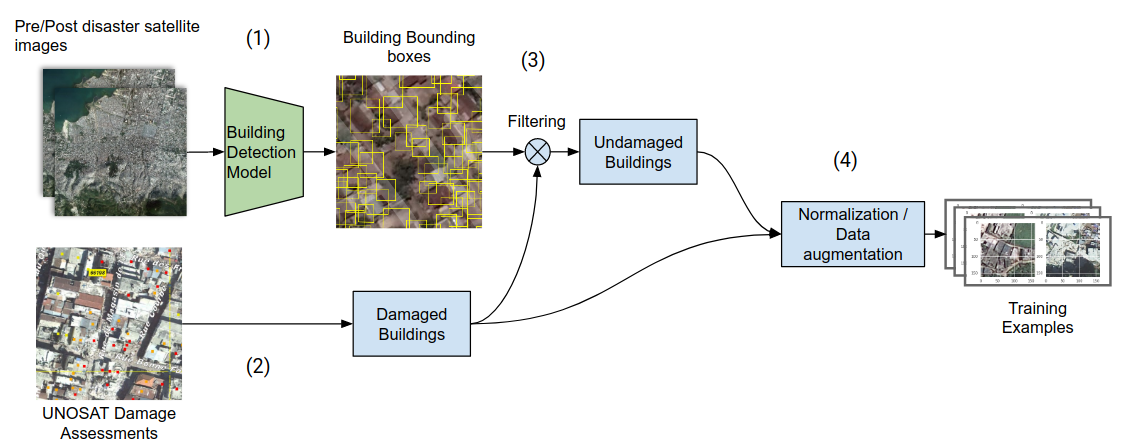}
\centering
\caption{The data generation pipeline: (1) Pre- and post-disaster satellite images are first passed through the building detection model to identify all buildings. (2) Damaged buildings are extracted from manual damage assessments of the region provided by UNOSAT. (3) Negative examples are obtained by removing the buildings tagged as damaged from all detected buildings. (4) Damaged and undamaged examples are normalized, and data augmentation is applied.}
\label{fig:data_pipeline}
\end{figure}

At the time of this work, there did not exist a comprehensive, multi-disaster data set for training a satellite building damage assessment model.
We created our own data set spanning three different disasters: the 2010 Haiti earthquake, 2017 Mexico City earthquake, and the 2018 Indonesia earthquake.
The data generation process consists of four steps, as shown in figure \ref{fig:data_pipeline} and described in detail below.

\begin{itemize}
\item \textbf{Obtain Satellite Images}
We collected satellite images of the affected regions before and after each disaster event.
Most of the satellite images we used came from DigitalGlobe's WorldView 2 and 3 satellites.
Some of the images are freely available from DigitalGlobe's FirstLook database.
For the Haiti Earthquake, we used candid flyover images provided by the National Oceanic and Atmosphere Administration.
We resampled all images to 0.3 meter resolution to maintain consistency of pixel scales.

We intentionally kept data-cleaning and pre-processing to a minimum, to avoid having to perform these labor-intensive operations in a real disaster-response scenario.
The only preprocessing we performed was a standard histogram equalization \cite{computer_vision} to normalize the range of pixel intensity values across different satellite images.

\item \textbf{Identify Damaged Buildings}
We used building damage assessments provided by UNOSAT, the operational satellite applications programme of the United Nations Institute for Training and Research (UNITAR), as positive labels for our training examples.
UNOSAT has performed manual post-disaster damage assessments for each of the disasters we target.
We downloaded these datasets from the Humanitarian Data Exchange website \cite{hdx}.

UNOSAT damage assessments use a 5-level scale to grade building damage - "No Damage", "Possible Damage", "Moderate Damage", "Severe Damage", and "Destroyed".
However, the assessments were noisy and the gradings were sometimes inconsistent across different datasets.
To minimize labeling noise, we group the "Severe Damage" and "Destroyed" labels into a single "Damaged" class, and train our model to distinguish between damaged buildings and all other buildings.

\item \textbf{Identify Undamaged Buildings}
Most of the UNOSAT damage assessments only labeled the positions of damaged buildings, so negative examples were not readily available.
To generate examples of undamaged buildings, we first used a building detection ML model to identify all buildings in the damage assessment area, and then filtered out all buildings that were marked by UNOSAT analysts as damaged.
This approach allowed us to generate a large number of negative examples for each disaster without requiring slow manual annotation.

The building detection model uses a Faster-RCNN \cite{faster_rcnn} architecture and was trained on ~80k human-generated labels and 4 million lower quality auto-generated labels.
At the confidence threshold of 0.5, precision is 0.64, recall is 0.75.
We use a standard non-maximal suppression algorithm to de-duplicate the model's output.
Figure \ref{fig:haiti_building_detection} in the appendix shows a sample of the model's output when run on the Haiti pre-disaster image.

\item \textbf{Sample Image Patches}
The final step in the data generation pipeline is to create individual training examples by sampling small crops around damaged and undamaged building centers.
We use Google Earth Engine \cite{earth_engine} to spatially join building damage labels with satellite images to produce the crops.
Each example in our dataset contains a 6 channel, 161 x 161 image crop centered on the building of interest.
The 6 channels is a concatenation of the RGB channels of the pre- and post-disaster images.
The intensity values range from 0-1.0.
See Figure \ref{fig:before_after} in the appendix for training examples from each dataset.
\end{itemize}

Table \ref{table:dataset} shows the numbers of examples we collected for each disaster.
\begin{table}[h!]
\centering
\caption{Dataset details}
\begin{tabular}{|p{5cm} p{3cm} p{2cm} p{2cm}|} 
\hline
Event & Location & Num. pos. & Num. neg. \\ 
\hline
Haiti 2010 Earthquake & Port-au-Prince & 31489 & 37214 \\
Mexico City 2017 Earthquake & Cuernavaca & 1494 & 2940 \\
Indonesia 2018 & Gumantar & 1274 & 1057 \\
\hline
\end{tabular}
\label{table:dataset}
\end{table}

\section{\label{sec:s3}Model Architecture Comparisons}

\begin{figure}[ht]
\centering
\hspace*{-1.2em}
\subfloat[CC]
{ \includegraphics[width=1.1in]{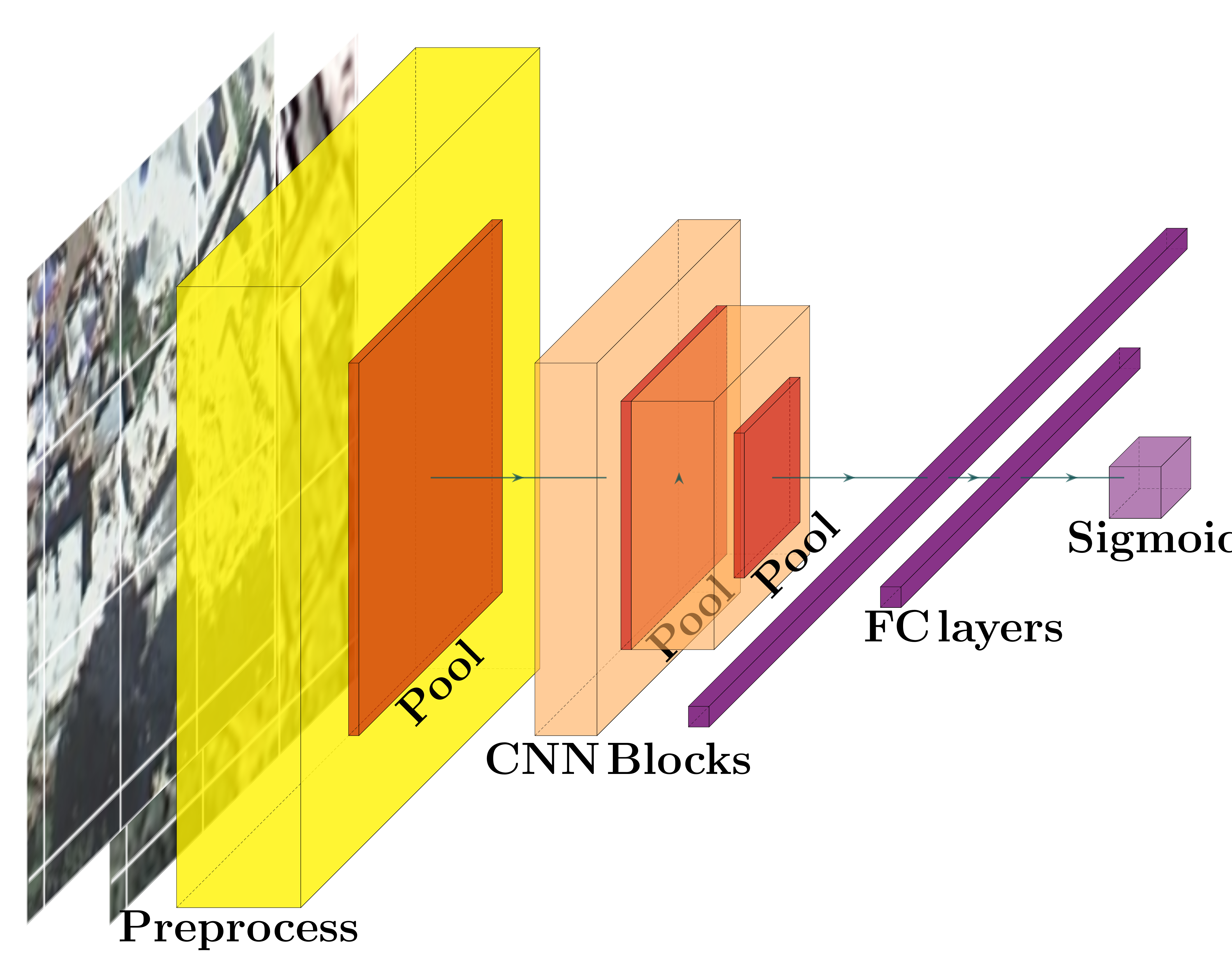}
}
\subfloat[PO]
{ \includegraphics[width=1.0in]{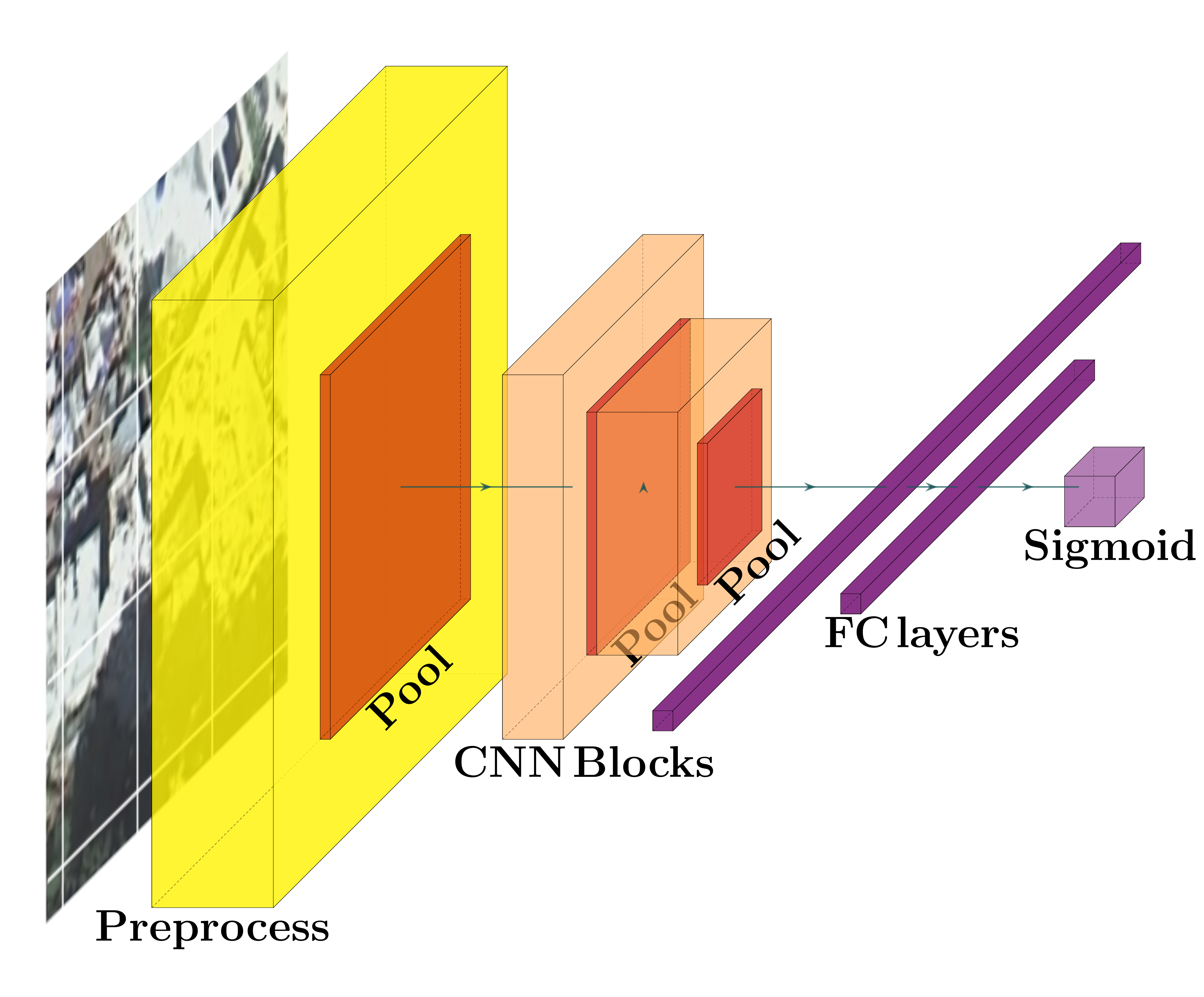}
}
\subfloat[TTC]
{ \includegraphics[width=1.1in]{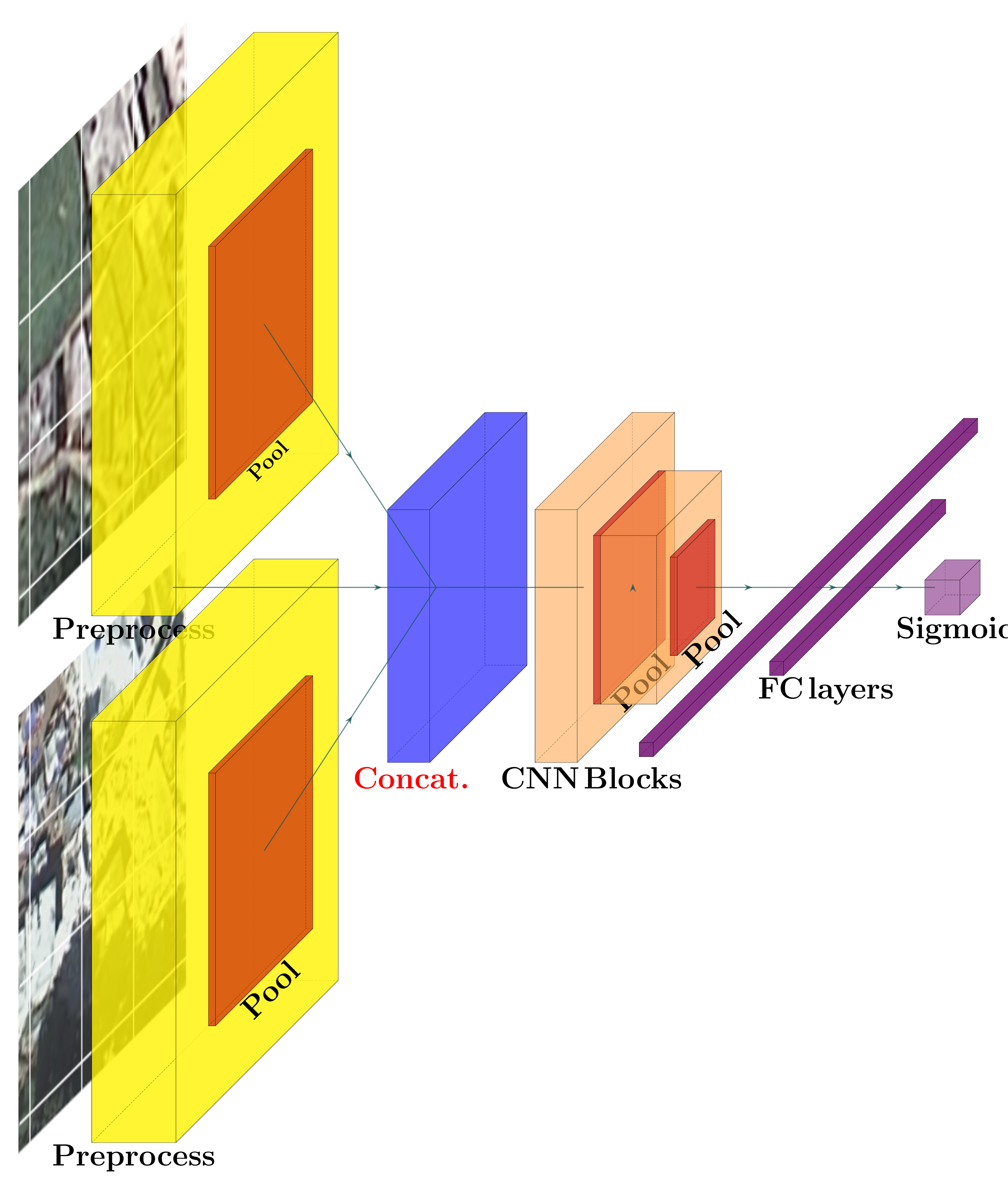}
}
\subfloat[TTS]
{ \includegraphics[width=1.1in]{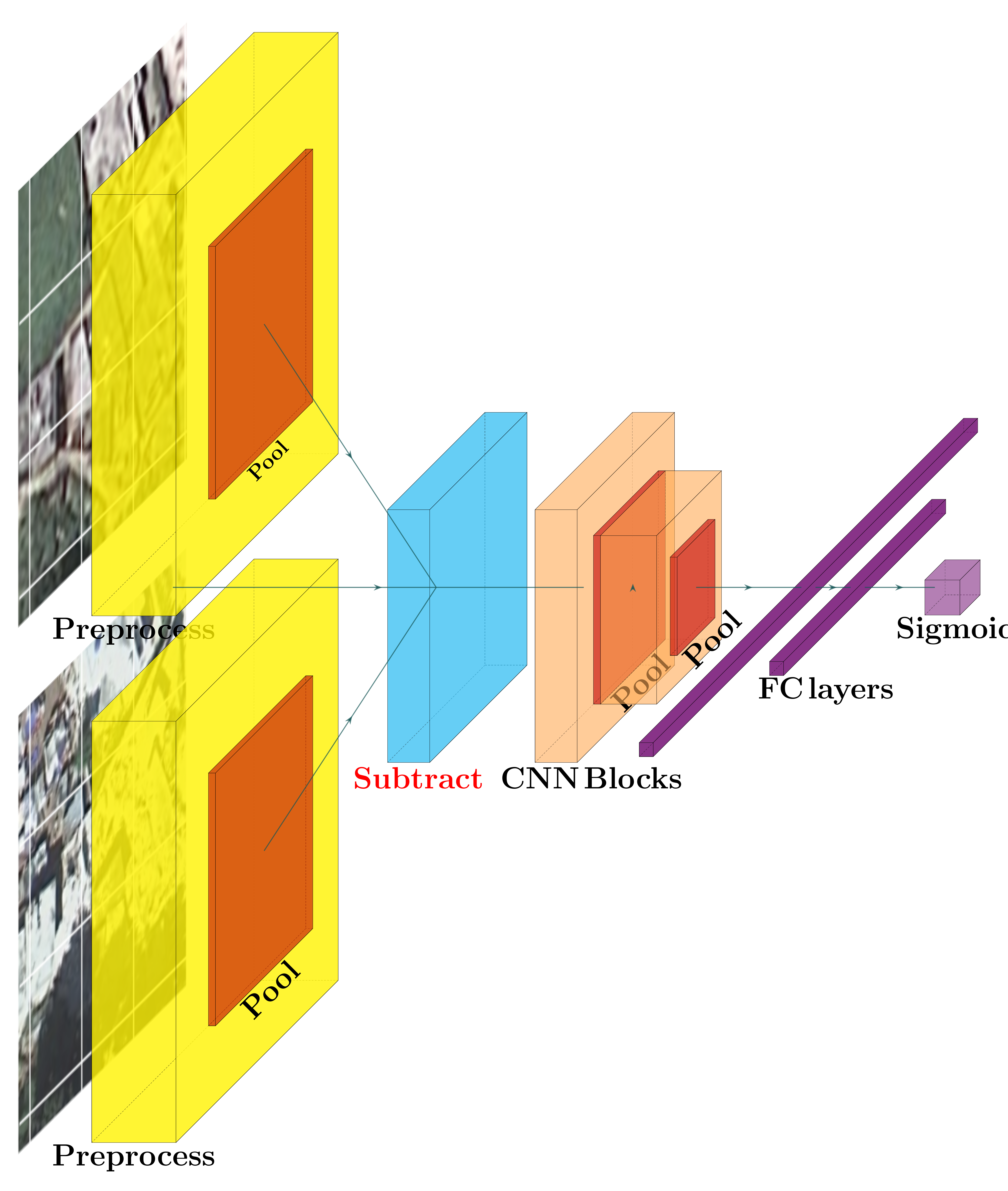}
}
\caption{
Model architecture variants.
(CC) and (PO) are single tower architectures.
(CC) concatenates the RGB channels of pre- and post-disaster images as input.
(PO) only uses the post-disaster image as input.
(TTC) and (TTS) are twin-tower architectures that process the pre- and post-disaster images separately using convolutional blocks before combining them and feeding into second half of the network.
(TTC) concatenates the outputs of the separate convolutional layers, while (TTS) subtracts the outputs.
The second half of all four architectures are identical: 1 convolutional block followed by 2 fully connected layers and a sigmoid output layer.}
\label{fig:models}
\end{figure}

We use convolutional neural networks (CNNs) as the basis of our models.
We experimented with four different CNN architectures, as shown in Figure\ref{fig:models}.
All four architectures follow the AlexNet architecture\cite{alexnet:2017}, which uses a sequence of convolutional layers followed by a sequence of fully connected layers and finally a sigmoid layer as output.
The difference in the four architectures is in how the input images are processed, as described below.

\begin{itemize}
\item \textbf{Concatenated Channel (CC) Model}
Concatenate the pre- and post-disaster images into a single 6-channel image.
We use this architecture as a baseline.
\item \textbf{Post-image Only (PO) Model} 
Only use the 3-channel post-disaster image as input.
This model loses the information from the pre-disaster image, but avoids problems such as misalignment and brightness differences in the pre- and post-disaster images.
\item \textbf{Twin-tower Concatenate (TTC) Model}
Preprocess the pre- and post-disaster images using separate convolutional feature extractors, then concatenate the extracted features along the channel dimension.
This architecture is designed to compare the pre- and post-disaster images based on abstract features extracted by the convolutional layers, instead of comparing pixels directly.
This makes the model more robust to non-uniformity in the pre- and post- images, such as misalignment.
\item \textbf{Twin-tower Substract (TTS) Model}
Same as TTC, except combine the extracted feature values by subtracting them element-wise instead of concatenating them.
This architecture is designed to more directly capture the differences in the pre- and post-disaster images, which is a good indicator of building damage.
\end{itemize}

We evaluated the performance of each architecture on the Haiti earthquake dataset using 5-fold cross-validation \cite{kfold:2011}.
We use the area under the RoC curve (AUC) as the primary metric of model performance evaluation because it is robust to class imbalance and more indicative of model quality compared to the conventional accuracy measure \cite{auc}.

Experiment results (Table \ref{table:model_comparisons}) shows that twin-tower models outperform single tower models, and the TTS model achieves the best performance with 0.8302 validation AUC.
The better performance of the twin-tower models indicates that useful information can be extracted by comparing buildings and their surroundings in the post-disaster images against those in the pre-disaster images.
It is interesting that the TTS model outperforms the TTC model.
Theoretically, the TTC model is more general and should be able to emulate the subtraction layer in the TTS model and achieve an equal or better AUC, if that is the best way to use the input features.
We suspect that this didn't happen in our experiments because the training set is too small and the TTC model is overfitting the data.
Another interesting result is that the PO model outperforms the CC model, which has strictly more information.
This suggests that simply concatenating the pre- and post-disaster images without first extracting high level features doesn't allow the model to compensate for differences between the images such as object misalignment, camera angle differences, etc.

Based on these results, we use the TTS model in all subsequent experiments.

\begin{table}[h!]
\centering
\caption{Performance comparison of different architectures on the Haiti dataset.}
\begin{tabular}{c   c} 
\hline
Architectures & AUC \\ [0.5ex] 
\hline
CC & 0.8008 $\pm$ 0.0033 \\
PO & 0.8030 $\pm$ 0.0064 \\
TTC & 0.8120 $\pm$ 0.0054 \\
TTS & 0.8302 $\pm$ 0.0056 \\
\hline
\end{tabular}
\label{table:model_comparisons}
\end{table}

\section{\label{sec:s4}Cross-Region Generalization}

For a damage detection model to be practically useful, it must be able to perform well in future disasters.
In other words, the model must generalize well to disasters it has not been trained on.
The typical way to improve generalization in ML models is to increase training data size and variation \cite{Goodfellow:2016:DL:3086952}.
This is a challenge in the damage assesment domain because there are only a small number of past disasters for which high resolution satellite imagery and manual damage assessments are available.
For example, even the upcoming xBD dataset \cite{xbd}, which is the most comprehensive dataset of this type, only has earthquake data for four distinct geographic regions.
This means that there is limited variation in building characteristics, lighting conditions, terrain types, satellite image quality and camera angles in the training data.
The lack of training data variability means that the model can easily overfit on the training data and perform poorly out-of-sample.

In this section, we evaluate how well the model can generalize in the face of limited training data variability.
We experimented with the following training and testing conditions:

\begin{enumerate}
    \item Train and test the model on the same dataset.
          This establishes a best-scenario baseline of model performance for other experiments.
    \item Train the model on Haiti dataset, test the model on each of the other datasets.
          This establishes the worst case where the model is trained on examples with minimum diversity, and will likely overfit on them.
    \item Train the model on two datasets, test on a third dataset.
          This allows the model to learn from a more diverse set of examples, which reduces overfitting.
    \item Train the model on two datasets and one fold of the third dataset, test on the rest of the third dataset.
          This scenario reflects the possibility of obtaining a small amount of labeled data from manual annotators working on a disaster in real time.
\end{enumerate}

For the first condition, we partition the training and test sets into 10 equally sized folds, and use 8 folds for training and 2 folds for validation.
Instead of randomly assigning buildings to folds, we assign each fold an interval of longitudes, and assign all buildings in that interval to the fold.
This minimizes the chance that the model will see buildings from the test set in the periphery of buildings from the training set, a form of information leakage.
To reduce overfitting, we normalize all examples with histogram equalization and apply random augmentation with color manipulation, flipping and rotation.

\begin{table}[h!]
\centering
\caption{Results of generalizability experiments.}
\begin{tabular}{| l l l l|}
\hline
Train datasets & Test datasets & AUC & Accuracy \\
\hline
Mexico                               & Mexico    & 0.79 & 0.71 \\
Haiti                                & Mexico    & 0.62 & 0.60 \\
Haiti + Indonesia                    & Mexico    & 0.73 & 0.68 \\
Haiti + Indonesia + $10\%$ of Mexico & $90\%$ of Mexico    & 0.76 & 0.72 \\
\hline
Indonesia                            & Indonesia & 0.86 & 0.78 \\
Haiti                                & Indonesia & 0.63 & 0.60 \\
Haiti + Mexico                       & Indonesia & 0.73 & 0.67 \\
Haiti + Mexico + $10\%$ of Indonesia & $90\%$ of Indonesia & 0.80 & 0.70 \\
\hline
\end{tabular}
\label{table:gen_exps}
\end{table}

We only use the Haiti dataset as training data because it’s significantly larger than the other two.
We use AUC as the primary performance measure, but also report model accuracy for a better comparison with prior work.
Threshold values for computing model accuracy were obtained by performing a grid-search on the training data set.

The experiment results are shown in Table \ref{table:gen_exps}).
As expected, the experiments where training and test examples come from the same dataset achieve the highest AUCs, because of higher consistency in building characteristics, image quality and manual annotation standard.
While AUC is poor when the model is trained on only one dataset and tested on another, we see that it improves when trained on two datasets.
This suggests that region variability in the training data is important, even if the variability doesn't come from the same region as the test examples.
Finally, the result of experiment 4 suggests that the best cross-region results can be obtained if human annotators can manually assess building damage in a small neighborhood to fine-tune our model, after which the model can be used to assess the rest of the larger area.

\section{\label{sec:s5}Conclusion}
In this paper, we described a method to build convolutional neural networks that automatically detect damaged buildings in satellite images.
We introduced a novel way to generate large numbers of negative training examples automatically in our data generation pipeline.
We experimented with multiple model architectures and found the "two tower subtract" variant to perform the best at this task.
Finally, we empirically showed that the model can generalize well to new regions and disasters if it is fine-tuned on a small set of examples from that region.

For future work, we plan to investigate additional disaster types, especially hurricanes and armed conflicts.
We also plan to investigate techniques to make the model more robust to data flaws.
For example, we can introduce random translations in the training images to make the model more robust to misalignment between pre- and post-disaster satellite images.

\bibliographystyle{plain}
\bibliography{references}

\newpage

\section*{Acknowledgements}
This work is a collaboration between Google AI and the United Nations World Food Programme (WFP) Innovation Accelerator.
The WFP Innovation Accelerator identifies, supports and scales high-potential solutions to hunger worldwide.
We support WFP innovators and external start-ups and companies through financial support, access to a network of experts and a global field reach.
We believe the way forward in the fight against hunger is not necessarily in building grand plans, but identifying and testing solutions in an agile way.
The Innovation Accelerator is a space where the world can find out what works and what doesn’t in addressing hunger - a place where we can be bold, and fail as well as succeed.

\section*{\label{sec:appendix}Appendix A}

\begin{figure}[H]
\centering
\subfloat[]
{ \includegraphics[width=2.5in]{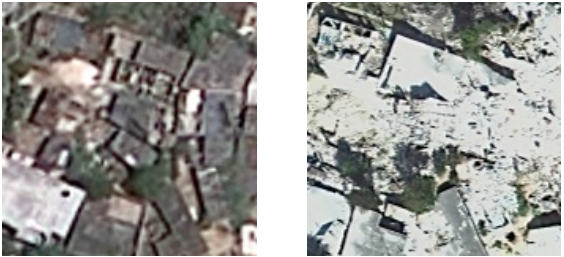}
}
\subfloat[]
{ \includegraphics[width=2.5in]{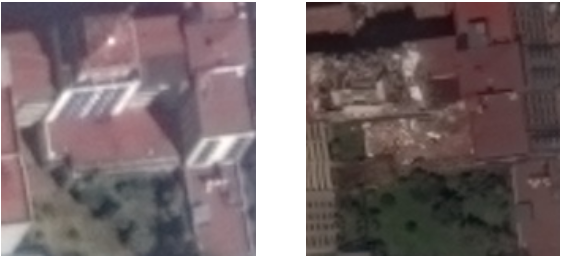}
}
\newline
\subfloat[]
{ \includegraphics[width=2.5in]{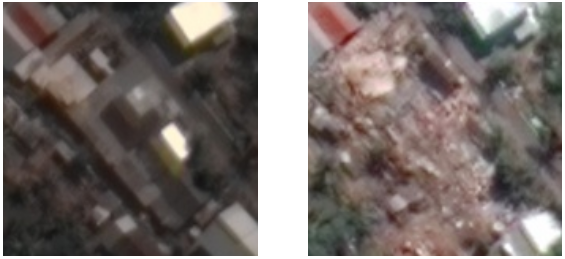}
}
\caption{Pre-disaster (left) and post-disaster (right) images from (a) the 2010 Haiti earthquake, (b) 2017 Mexico City earthquake, and (c) 2018 Indonesia earthquake.}
\label{fig:before_after}
\end{figure}

\begin{figure}[H]
\includegraphics[width=0.6\textwidth]{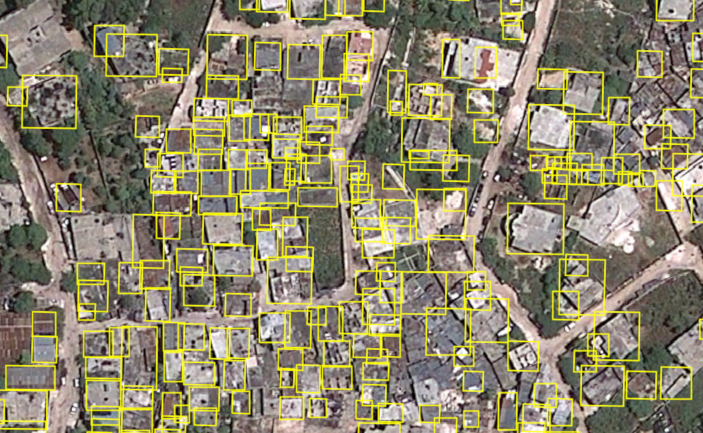}
\centering
\caption{Sample of building detection model output on Haiti}
\label{fig:haiti_building_detection}
\end{figure}

\end{document}